# Deep Learning in Multiple Multistep Time Series Prediction

Chuanyun (Clara) Zang

## Abstract

The project aims to research on combining deep learning specifically Long-Short Memory (LSTM) and basic statistics in multiple multistep time series prediction. LSTM can dive into all the pages and learn the general trends of variation in a large scope, while the well selected medians for each page can keep the special seasonality of different pages so that the future trend will not fluctuate too much from the reality. A recent Kaggle competition on 145K Web Traffic Time Series Forecasting [1] is used to thoroughly illustrate and test this idea.

# Definition

**Domain Background**

From stock market prices, weather forecasting to speech and image recognition, real world data is usually wildly collected by taking account of time. This kind of data is known as time series. Different fields encapsulate different problems depending on the purpose of data, for example finance market is interested in the stock prices for the future days while a robot system may need to recognize the next visual signal. Therefore, the time series analysis can range from classification, segmentation, to anomaly detection and prediction.

Time series analysis is attracting by its challenge in feature extraction and its openness to technique exploration. Traditional statistical method like ARIMA and machine learning algorithms like deep neural networks can find a place in this field. In this paper, a combination of some basic statistics and recurrent neural network (RNN) is applied to multistep time series prediction.

**Problem Statement**

The problem is from a Kaggle competition – Web Traffic Time Series Forecasting [1]. Given approximately 145K Wikipedia articles with their historical daily views, starting from July $1^{st}$, 2015 to December $31^{st}$, 2016, the goal is to forecast future daily web traffic, from January $1^{st}$, 2017 up until March $1^{st}$, 2017 for each article. For example, we can simply use today's views to predict tomorrow's views for all these Wikipedia articles, and the solution can be measured by comparing the difference between prediction and true values. The challenge here is that there are approximately 145K time series instead of only one, and the fact that daily views follow a random walk and hence the future daily views are hard to be predicted. We combine deep learning and statistics on this problem which showed power in tackling it.

---

**Datasets and Inputs**

The datasets used are data from Stage 1 of the Kaggle competition (see [1] -> Data). The training dataset consists of approximately 145K time series. Each time series contains a Page name and a sequence of daily views for the corresponding Page, starting from 2015-07-01 to 2016-12-31. The Page name is made of article name, the Wikipedia project like 'en.wikipedia.org', type of access such as desktop and type of agent like spider, which is in the format of 'name_project_access_agent'. For values in each time series of daily views, there is no difference between zero values and missing values, so a missing value may mean the traffic was zero or just that the data is not available for that day. Now since the stage 1 of the competition is over, I will also use another dataset containing the daily views for January 1$^{st}$, 2017 up until March 1$^{st}$, 2017 as the answer key to evaluate the model prediction performance.

| | Page | 2015-07-01 | 2015-07-02 | 2015-07-03 | 2015-07-04 | 2015-07-05 | 2015-07-06 | 2015-07-07 | 2015-07-08 | 2015-07-09 | ... | 2016-12-22 | 2016-12-23 | 2016-12-24 | 2016-12-25 | 2016-12-26 | 2016-12-27 | 2016-12-28 | 2016-12-29 | 2016-12-30 | 2016-12-31 |
|---|---|---|---|---|---|---|---|---|---|---|---|---|---|---|---|---|---|---|---|---|---|
| 0 | 2NE1_zh.wikipedia.org_all-access_spider | 18.0 | 11.0 | 5.0 | 13.0 | 14.0 | 9.0 | 9.0 | 22.0 | 26.0 | ... | 32.0 | 63.0 | 15.0 | 26.0 | 14.0 | 20.0 | 22.0 | 19.0 | 18.0 | 20.0 |
| 1 | 2PM_zh.wikipedia.org_all-access_spider | 11.0 | 14.0 | 15.0 | 18.0 | 11.0 | 13.0 | 22.0 | 11.0 | 10.0 | ... | 17.0 | 42.0 | 28.0 | 15.0 | 9.0 | 30.0 | 52.0 | 45.0 | 26.0 | 20.0 |
| 2 | 3C_zh.wikipedia.org_all-access_spider | 1.0 | 0.0 | 1.0 | 1.0 | 0.0 | 4.0 | 0.0 | 3.0 | 4.0 | ... | 3.0 | 1.0 | 1.0 | 7.0 | 4.0 | 4.0 | 6.0 | 3.0 | 4.0 | 17.0 |
| 3 | 4minute_zh.wikipedia.org_all-access_spider | 35.0 | 13.0 | 10.0 | 94.0 | 4.0 | 26.0 | 14.0 | 9.0 | 11.0 | ... | 32.0 | 10.0 | 26.0 | 27.0 | 16.0 | 11.0 | 17.0 | 19.0 | 10.0 | 11.0 |
| 4 | 52_Hz_I_Love_You_zh.wikipedia.org_all-access_spider | NaN | NaN | NaN | NaN | NaN | NaN | NaN | NaN | NaN | ... | 48.0 | 9.0 | 25.0 | 13.0 | 3.0 | 11.0 | 27.0 | 13.0 | 36.0 | 10.0 |

Table 1. Data Sample

**Evaluation Metrics**

In the Kaggle competition, **SMAPE** between predicted values and true values is used as the evaluation metric. SMAPE is short for Symmetric Mean Absolute Percentage Error, which is an accuracy measure based on percentage errors. The formula of SMAPE is

$$\text{SMAPE} = \frac{200\%}{n} \sum_{t=1}^{n} \frac{|y_t - \hat{y}_t|}{|y_t| + |\hat{y}_t|}$$

where n is the number of observations, $y_t$ is the true value and $\hat{y}_t$ is the predicted value where t = 1, ..., n. SMAPE has a lower bound 0 and an upper bound 200%. One problem of SMAPE is that if the true value and predicted value are both 0, the formula returns error. Hence in practice, we define that if $|y_t| + |\hat{y}_t| = 0$, then SMAPE = 0; otherwise SMAPE is calculated by the above formula. Because our data contains lots of extreme values, i.e. outliers, SMAPE does not return large error for them by taking percentage, which SMAPE is good metric for testing our final results.

In the training process, I also use another metric Mean Absolute Error (**MAE**) which is a quality to measure how close predictions are to the actual values, and defined as

$$\text{MAE} = \frac{1}{n}\sum_{t=1}^{n}|y_t - \hat{y}_t|$$

where n is the number of observations, $y_t$ is the true value and $\hat{y}_t$ is the predicted value where t = 1, …, n. MAE is used as a loss function during training because the data is log transformed and error $\log(y_t) - \log(\hat{y}_t) = \log\left(\frac{y_t}{\hat{y}_t}\right)$ should be close to SMAPE in functionality in the manner of using percentage of outputs.

## Analysis

### Data Exploration

The basic information of the data: there are 145,063 rows and each row contains a column Page (in 'name_project_access_agent' format) and a sequence of 550 daily views from 2015-07-01 to 2016-12-31.

```
Dataframe.info()
<class 'pandas.core.frame.DataFrame'>
RangeIndex: 145063 entries, 0 to 145062
Columns: 551 entries, Page to 2016-12-31
dtypes: float64(550), object(1)
```

There are lots of missing values by page and by date. From page aspect, most pages have the daily views for all days; but there are some exceptions, for example 652 pages have no daily views recorded at all. From date aspect, each day has some missing views for at least 3,189 pages and up to 20,816 pages.

```
Missing values by row:                    Missing values by date:
mean       42.691320                      mean     11259.874545
std       115.804572                      std       5275.772841
min         0.000000                      min       3189.000000
25%         0.000000                      25%       6614.500000
50%         0.000000                      50%      10560.500000
75%         0.000000                      75%      15792.500000
max       550.000000                      max      20816.000000
```

There are outliers both along page and along date. For each page, there are extreme values at different time points. See Table 2 for the summary and Figure 1 for the plot with high spikes.

| Page  | 0     | 1     | 2     | 3     | 4     | 5     | 6     | 7     | 8     | 9     | ... | 145053 | 145054 | 145055 | 145056 | 145057 | 145058 | 145059 | 145060 | 145061 | 145062 |
|-------|-------|-------|-------|-------|-------|-------|-------|-------|-------|-------|-----|--------|--------|--------|--------|--------|--------|--------|--------|--------|--------|
| count | 550   | 550   | 550   | 550   | 550   | 550   | 550   | 550   | 550   | 550   | ... | 550    | 550    | 550    | 550    | 550    | 550    | 550    | 550    | 550    | 550    |
| mean  | 21.76 | 25.39 | 5.20  | 17.13 | 4.84  | 16.41 | 6.62  | 49.65 | 35.01 | 22.33 | ... | 1.20   | 0.90   | 0.31   | 1.18   | 0.04   | 0.10   | 0.00   | 0.00   | 0.00   | 0.00   |
| std   | 29.39 | 33.98 | 13.51 | 19.34 | 17.32 | 15.67 | 14.98 | 50.66 | 21.12 | 9.29  | ... | 6.57   | 4.11   | 3.41   | 11.03  | 0.90   | 1.05   | 0.00   | 0.00   | 0.00   | 0.00   |
| min   | 3     | 2     | 0     | 1     | 0     | 1     | 0     | 1     | 5     | 2     | ... | 0      | 0      | 0      | 0      | 0      | 0      | 0      | 0      | 0      | 0      |
| 25%   | 11    | 13    | 2     | 9     | 0     | 9     | 0     | 18    | 24    | 16    | ... | 0      | 0      | 0      | 0      | 0      | 0      | 0      | 0      | 0      | 0      |
| 50%   | 16    | 17    | 4     | 13    | 0     | 14    | 0     | 35    | 31    | 22    | ... | 0      | 0      | 0      | 0      | 0      | 0      | 0      | 0      | 0      | 0      |
| 75%   | 22    | 27    | 5     | 19    | 5     | 19    | 8     | 64    | 41    | 28    | ... | 0      | 0      | 0      | 0      | 0      | 0      | 0      | 0      | 0      | 0      |
| max   | 490   | 621   | 210   | 303   | 234   | 213   | 121   | 520   | 203   | 62    | ... | 86     | 46     | 61     | 166    | 21     | 13     | 0      | 0      | 0      | 0      |

**Table 2**. Data Summary by Page

As aspect of date, there are also outliers. Each day different pages receive different views but the range is extremely large. For example, for the first date 2015-07-01, the range is between 0 to 2,038,124 with median only equal to 58. See Table 3 for the summary along date.

| | Page | 2015-07-01 | 2015-07-02 | 2015-07-03 | 2015-07-04 | 2015-07-05 | 2015-07-06 | 2015-07-07 | 2015-07-08 | 2015-07-09 | ... | 2016-12-22 | 2016-12-23 | 2016-12-24 | 2016-12-25 | 2016-12-26 | 2016-12-27 | 2016-12-28 | 2016-12-29 | 2016-12-30 | 2016-12-31 |
|---|---|---|---|---|---|---|---|---|---|---|---|---|---|---|---|---|---|---|---|---|---|
| count | 145063 | 1.450630e+05 | 1.450630e+05 | 1.450630e+05 | 1.450630e+05 | 1.450630e+05 | 1.450630e+05 | 1.450630e+05 | 1.450630e+05 | 1.450630e+05 | ... | 1.450630e+05 | 1.450630e+05 | 1.450630e+05 | 1.450630e+05 | 1.450630e+05 | 1.450630e+05 | 1.450630e+05 | 1.450630e+05 | 1.450630e+05 | 1.450630e+05 |
| unique | 145063 | NaN | NaN | NaN | NaN | NaN | NaN | NaN | NaN | NaN | ... | NaN | NaN | NaN | NaN | NaN | NaN | NaN | NaN | NaN | NaN |
| top | 蘇打綠_zh.wikipedia.org_all-access_spider | NaN | NaN | NaN | NaN | NaN | NaN | NaN | NaN | NaN | ... | NaN | NaN | NaN | NaN | NaN | NaN | NaN | NaN | NaN | NaN |
| freq | 1 | NaN | NaN | NaN | NaN | NaN | NaN | NaN | NaN | NaN | ... | NaN | NaN | NaN | NaN | NaN | NaN | NaN | NaN | NaN | NaN |
| mean | NaN | 1.024882e+03 | 1.031234e+03 | 9.731234e+02 | 1.003791e+03 | 1.044342e+03 | 1.108086e+03 | 1.062624e+03 | 1.026181e+03 | 1.030808e+03 | ... | 1.357067e+03 | 1.343449e+03 | 1.362474e+03 | 1.484413e+03 | 1.634242e+03 | 1.635483e+03 | 1.590915e+03 | 1.639885e+03 | 1.431160e+03 | 1.442972e+03 |
| std | NaN | 6.735340e+04 | 6.868551e+04 | 6.449417e+04 | 6.720994e+04 | 6.834087e+04 | 7.464303e+04 | 7.016081e+04 | 6.325117e+04 | 6.632223e+04 | ... | 8.460264e+04 | 7.636701e+04 | 8.384845e+04 | 8.638560e+04 | 9.661396e+04 | 9.113984e+04 | 9.064050e+04 | 8.894637e+04 | 8.052685e+04 | 8.766977e+04 |
| min | NaN | 0.000000e+00 | 0.000000e+00 | 0.000000e+00 | 0.000000e+00 | 0.000000e+00 | 0.000000e+00 | 0.000000e+00 | 0.000000e+00 | 0.000000e+00 | ... | 0.000000e+00 | 0.000000e+00 | 0.000000e+00 | 0.000000e+00 | 0.000000e+00 | 0.000000e+00 | 0.000000e+00 | 0.000000e+00 | 0.000000e+00 | 0.000000e+00 |
| 25% | NaN | 4.000000e+00 | 3.000000e+00 | 4.000000e+00 | 4.000000e+00 | 4.000000e+00 | 4.000000e+00 | 4.000000e+00 | 5.000000e+00 | 5.000000e+00 | ... | 1.800000e+01 | 1.900000e+01 | 1.800000e+01 | 1.800000e+01 | 1.800000e+01 | 1.900000e+01 | 2.000000e+01 | 2.000000e+01 | 1.900000e+01 | 1.800000e+01 |
| 50% | NaN | 5.800000e+01 | 5.700000e+01 | 5.400000e+01 | 5.500000e+01 | 5.900000e+01 | 5.900000e+01 | 6.100000e+01 | 6.300000e+01 | 6.100000e+01 | ... | 1.370000e+02 | 1.330000e+02 | 1.220000e+02 | 1.330000e+02 | 1.460000e+02 | 1.490000e+02 | 1.490000e+02 | 1.470000e+02 | 1.420000e+02 | 1.250000e+02 |
| 75% | NaN | 4.090000e+02 | 4.060000e+02 | 3.980000e+02 | 3.880000e+02 | 4.280000e+02 | 4.340000e+02 | 4.310000e+02 | 4.360000e+02 | 4.350000e+02 | ... | 5.850000e+02 | 5.770000e+02 | 5.510000e+02 | 6.040000e+02 | 6.340000e+02 | 6.450000e+02 | 6.300000e+02 | 6.260000e+02 | 6.120000e+02 | 5.420000e+02 |
| max | NaN | 2.038124e+07 | 2.075219e+07 | 1.957397e+07 | 2.043964e+07 | 2.077211e+07 | 2.254467e+07 | 2.121089e+07 | 1.910791e+07 | 1.999385e+07 | ... | 2.420108e+07 | 2.253925e+07 | 2.505662e+07 | 2.586575e+07 | 2.834288e+07 | 2.691699e+07 | 2.702505e+07 | 2.607382e+07 | 2.436397e+07 | 2.614954e+07 |

**Table 3**. Data Summary by Date

**Exploratory Visualization**

First of all, let's fill the missing with 0 and plot some time series. Different page has various views pattern along time. It also shows that there are spikes in the time series which are outliers that we need to handle before modeling.

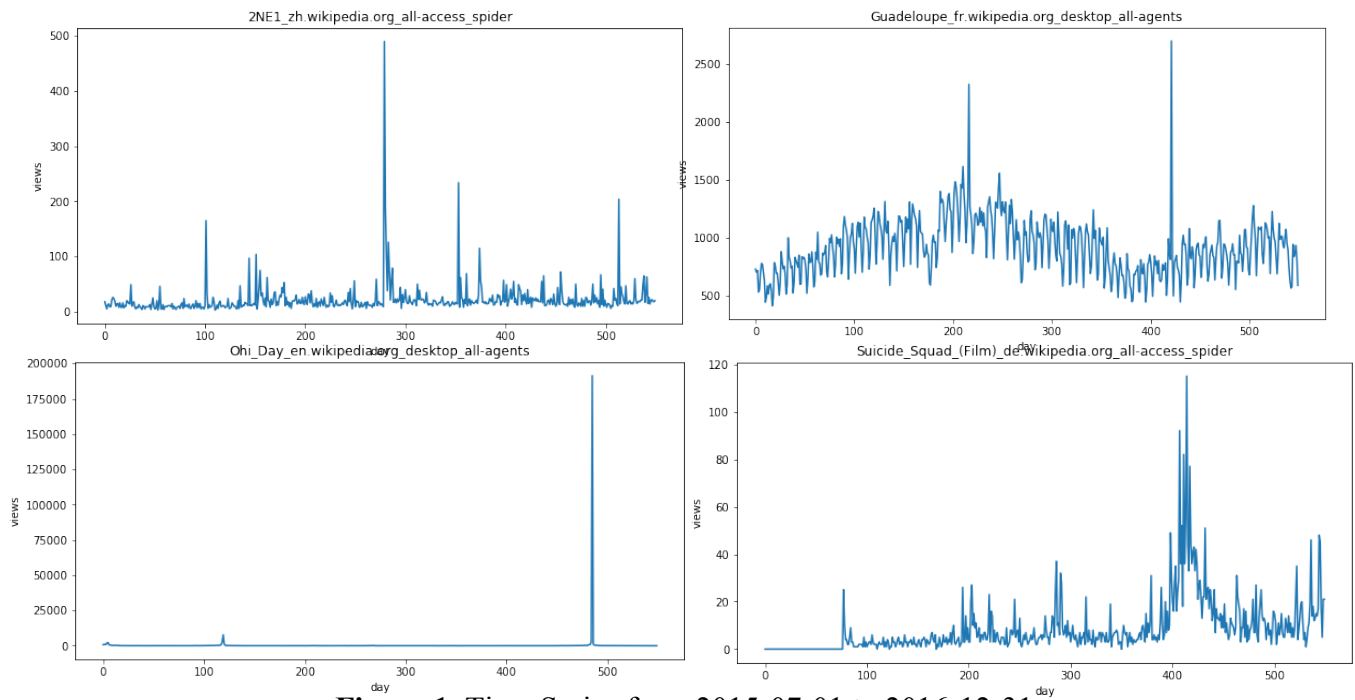

**Figure 1**. Time Series from 2015-07-01 to 2016-12-31

# Algorithms and Techniques

In this section, two techniques are discussed. One is a machine learning algorithm - long short term memory network, and the other is simply median of several aggregated medians.

1. Long-Short Term Memory Architecture

A Recurrent Neural Network (RNN) is a network with loops in them. Unrolling it, we can consider it as multiple copies of the same network (repeating module), and information is passed from one copy to the next one.

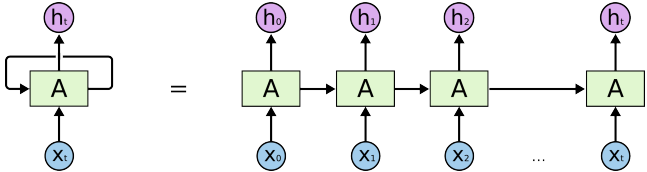

**Figure 2**: Chain-like Structure of RNNs ([3])

This chain-like nature makes RNNs a good tool for sequential data. While one big problem of RNNs is the long term dependency because of the vanishing/exploding gradient problem like other very deep Feedforward Neural Networks (FFNNs). Long-Short Term Memory (LSTM) networks, a special kind of RNNs, try to combat this problem by using gates and an explicitly defined memory cell. It was first introduced by Hochreit and Schmidhuber [2] in 1997 and generalized by many following work. The repeating module of a traditional LSTM contains 4 layers, interacting in a very special way.

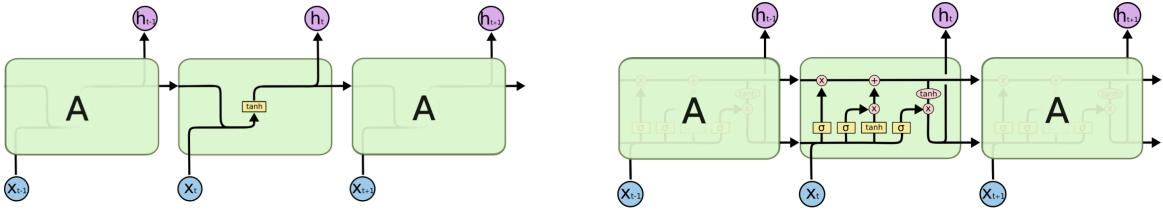

**Figure 3:** on the left is a simple RNN with one Hyperbolic Tangent layer in the repeating module; on the right is a traditional LSTM with 4 layers in its repeating module. ([3])

Since we have about 145K time series, it is not feasible to get different model for each time series, and traditional art of state models may not good at handling such huge different time series. The problem can be taken as a sequence-to-sequence prediction where one needs to forecast a sequence of future daily views by using a given sequence of previous daily views. Therefore, Long Short Term Memory with its chain-like structure is a good fit to it.

Before transforming data into a sequence to sequence analysis, we have to carefully handle the outliers in each time series. Besides, the range of the daily views for all 145K pages is hugely expanded, so we also need to properly scale the data before fitting into models. The details about data preprocessing will be documented in the later section. The Keras API will be used to build the Long-Short Term Memory network architecture with parameters like number of neurons, number of epochs, batch size tuned.

2. Median of Median

Due to the outliers (high spikes in Figure 1), I choose median to get a relevantly stable statistic for each time series. Further because of the variation along the 550 days, median of all 550 daily views is not a reasonable estimate. For example, some wiki page for a new movie only have views after there is any news about it, while some pages for events like regular games, holiday and so on get high hit in web traffic during some specific time in a year. Hence median of different period can capture different patterns or information about the time series.

In the later implementation, median of values in previous week is used to evaluated the most recent views; medians of values in previous 3 weeks, 9 weeks (which are 63 days, close to the 60-day range for our forecasting purpose), 1 year grouped by weekday respectively are used to capture the weekly patterns from a short to long term; median of the identical month in previous year, i.e. January and February 2016, is used to represent some potential yearly patterns. Finally taking median of all these medians will give a not too bad estimate of future daily views.

**Benchmark Model**

Since the prediction range is 60 days, the benchmark model I choose is to use the median of values of the previous 60 days as prediction of web views for the next 60 days for each Page. As stated in last section, median is not affected by outliers in the time series. With more than 145K sequences, it turned to be an easy and effective measurement in the predictions.

| Dataset | SMAPE |
| --- | --- |
| y_train | 45.53 |
| y_validate | 47.96 |
| y_test | 46.51 |

Table 4. SMAPE of benchmark model on different targets

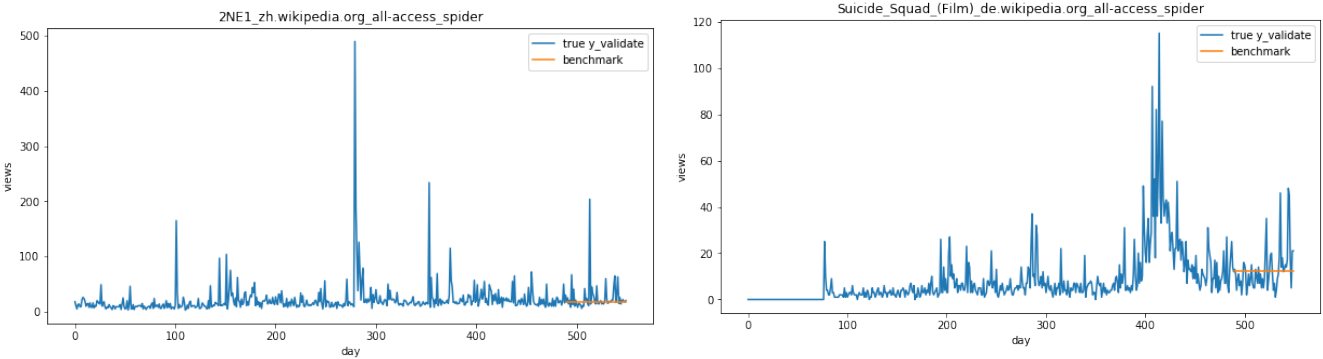

**Figure 4.** The prediction of benchmark on validating data set

# Methodology

**Data Preprocessing**

The range of daily views are huge, so necessary data transformation like log-scale, max-min scale or standardization is taken before modeling. We also need to appropriately define training set, validating set and testing sets. In this case, we use sequence of 60 values as target since we need to predict values for the next 60 days.

In details, the following preprocessing is taken.

1. Missing values. Fill missing values with 0. Another alternative mutation is to use the neighbors of it to evaluate.

2. Save true values y_true and y_validate for later performance checking, i.e. take the last 60 daily views as y_validate, and the 60 days before it as y_train.

3. Logarithmic transformation to deal with outliers. Since there are many 0's, log1p transformation is taken, i.e. $x' = \log(x+1)$. Alternative way to handle outliers that I tried is to shrink all daily views in the range (Q1-1.5*IQR, Q3+1.5*IQR) where Q1 is the 1st quartile, Q3 is the 3rd quartile and IQR = Q3-Q1.

4. Define X_train/y_train, X_validate/y_validate and X_test. In order to validate the model, I take the last 60 days as y_validate and the previous days except the first 60 days as X_validate. The choice of X_validate is to keep the same length as X_train which will be all days except the last 120 days, and y_train will be the 60 days before y_validate. In this case, there is no overlap between y_train and y_validate. So finally, X_test is all days except the first 120 days, which has the same length as X_train and X_validate, while y_test is unknow and represent views for the so-called future 60 days.

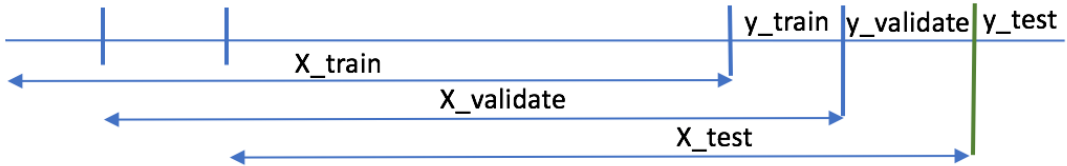

**Figure 5**. Defining X's and y's

5. Normalization X's. MinMaxScaler form the sklearn library is applied as follows. It scales the values between 0 and 1, and an alternative range is to reduce them between -1 and 1.

$$X\_norm = \frac{X_i - X_{min}}{X_{max} - X_{min}}$$

**Implementation**

In this section, two techniques are discussed. One is LSTM with its implementation in Keras and the other is the simply selected medians, followed by the ensemble of these two results.

1. LSTM Implementation

After logarithmic transformation and Max-Min scaling, we first reshape X's from a 2D array [samples, features] to 3D array [samples, timesteps, features] as required by LSTM network. Here we take timesteps as 1.

We design a simple LSTM network by using 1 hidden layer with 256 neurons followed by a dropout layer which is a regularization technique by randomly selecting a proportion of neuron to ignore, and then an output layer with linear activation and 60 output values. The network will be complied by MAE as loss function and RMSprop as optimizer.

We train the LSTM model on (X_train, y_train) and validate the model performance on (X_validate, y_validate) for each epoch. Callbacks function in keras is used to save the best shot on (X_validate, y_validate).

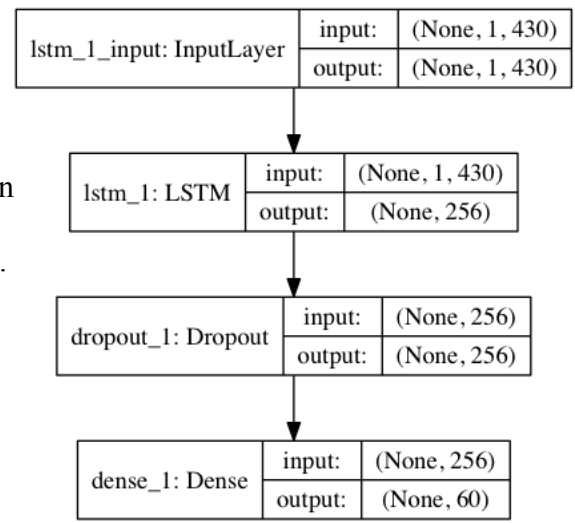

**Figure 5**. LSTM Network Architecture

When training the model, we can use batch size which is number of samples that are going to the be propagated through the network. It will help the networks run faster but the training accuracy is not as good as using the whole dataset. So I did not define batch size in this implementation. As to the number of epochs, several options are tried and 10 is chosen since the accuracy is not improved on (X_validate, y_validate) after several epochs. During the training process, each epoch takes about 80 seconds.

Keras library in python makes the implementation of neural networks easier with only several lines. But we still need to tune the hyperparameters. Here an approach is to use grid search. Due to the long running time, I did not try it. If we run it on GPU, grid search is better. Instead, I tested some hyperparameters manually. For example, number of neurons = 256 is better than 128. Dropout rate = 0.3 is better than 0.5. when using 25 or 35 epochs, the loss function stopped improving after around 10 epochs.

Choosing the proper loss function and optimization is also import. For loss function, I also tried Mean Square Error, SMAPE and Mean Absolute Percentage Error (MAPE), they did not outperform MAE when testing the final prediction of views. As documented in keras, RMSProp performs well for Recurrent Neural Networks.

SMAPE between true y_validate and predicted y_validate from LSTM models is 45.9.

2. Medians

As stated in earlier section, several medians are taken in different time ranges.

a) For each page, the median of views in the last week is taken.
b) For each page, the median of views in last 3 weeks grouped by weekday is taken.
c) For each page, the median of views in last 9 weeks grouped by weekday is taken.
d) For each page, the median of views during last 365 days (1 year) grouped by weekday is taken.
e) For each page, the median of views during 2016-01-01 to 2016-03-01 grouped by weekday is taken.

For this approach, I also tried adding median of other time range lie 6 weeks, 12 weeks, etc., but the final results were getting worse. How about other time range? I believe there is still a pace to improve it, but this is not the main focus of our methodology in general.

Finally, we join all the 5 medians by page and weekday, and then merge it to the LSTM results. The median of all these predictions will be the final predicted results.

**Refinement**

Tuning hyperparameters like number of neurons in LSTM and changing loss function will affect the performance. Besides that, to fully utilize the data set, I trained another LSTM model on (X_validate, y_validate). For this training process, the last 5% of (X_validate, y_validate) is used to validate the model performance.

In order to test the performance, we use the second model to predict on X_train. SMAPE between true y_train and predicted y_train is 45.4.

Taking average of predictions from these two LSTM models, one trained on (X_train, y_train) and the other trained on (X_validate, y_validate), as the final prediction of LSTM model.

## Results

**Model Evaluation and Validation**

Generalize and apply LSTM models on X_test. The $1^{st}$ model trained on (X_train, y_train) returned SMAPE between true y_test and predicted y_test as 46.15, and the $2^{nd}$ model trained on (X_validate, y_validate) returned SMAPE between true y_test and predicted y_test as 46.86. And the final LSTM model returned SMAPE as 45.9.

As to the medians, if we only take median of 5 medians as the prediction of y_test, SMAPE between the true and predicted values is 44.29. It is better than the prediction from LSTM model. There is no surprise since median is a stable statistic for each page level while LSTM network here is trying to capture the patterns of all pages at once.

Taking median of the 5 medians and the prediction from LSTM will further reduce the SMAPE value to 43.59.

Overall, the model generalized well to new dataset with a smaller SMAPE compared to the benchmark model. The use of medians protects the model from large fluctuations and it is robust to outliers. LSTM dives into all the pages and tries to learn the general processing trends so it helps providing a direction, while the mere medians for each page can capture the specialties of different pages so that the future trend won't go too far from the reality.

**Justification**

Compared to SMAPE between y_test and prediction from benchmark is 46.5, so overall our models with SMAPE=43.59 outperform it. How about the performance for each page? After taking a closer look at the results, it shows that LSTM works better for some pages while median of weekly medians work better for others. Even for the same page, LSTM works well for some time period while medians perform more stable for other period.

For example, for the following page, LSTM can capture the fluctuation in some days while median of medians draw it closer to the true values when the views are small.

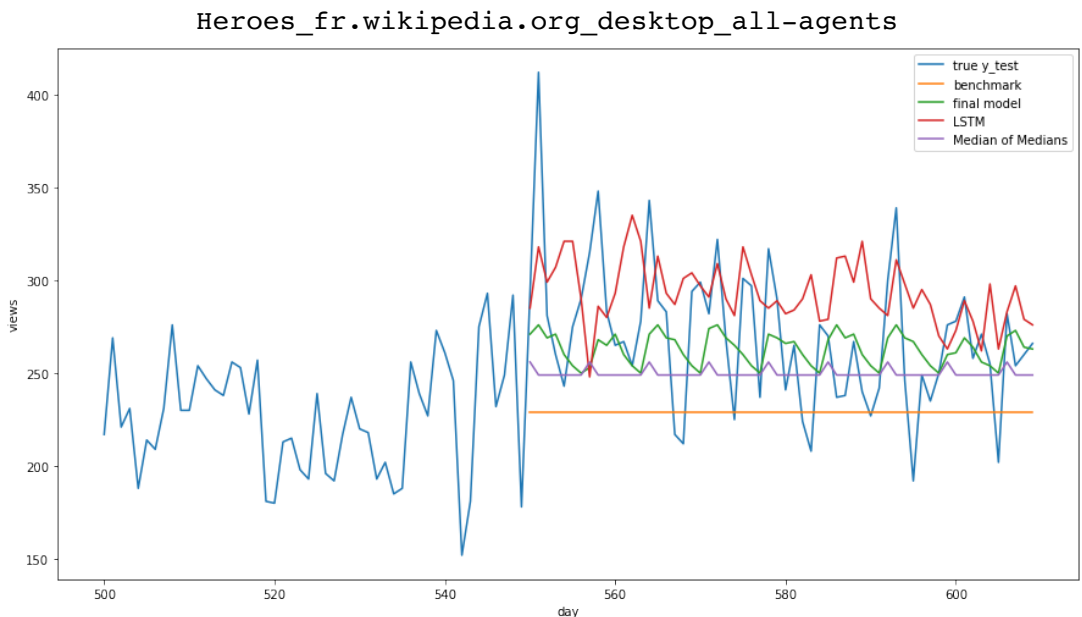

**Figure 6**. True values and predictions from different methods

The following page has a non-stationary trend of daily views. Median od medians cannot capture the trend as it decreases or increase a lot. While LSTM is good at catching the trend in a short term, and cannot make the quick jump at the end.

`Template:Syrian_Civil_War_detailed_map_en.wikipedia.org_desktop_all-agents`

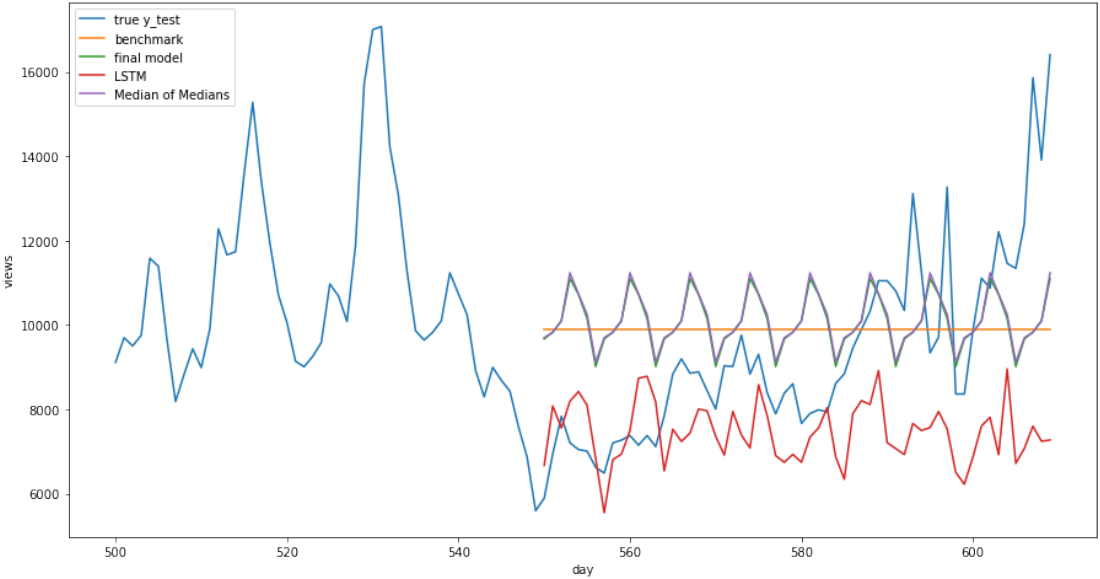

**Figure 7**. True values and predictions from different methods

The following page has a weekly seasonality with an increasing trend. A stable median of medians captures the seasonality but LSTM captures the variation in the seasonality. So the final model of combining these two performs better in general.

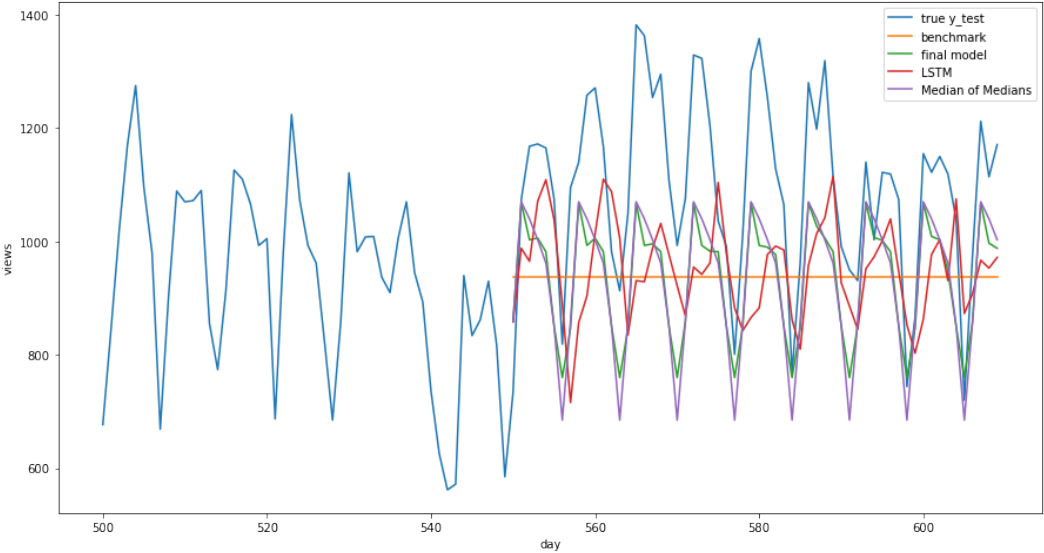

**Figure 8**. True values and predictions from different methods

The following page has very few daily views with mostly less than 5 views. The median of the 5 selected medians is almost a constant. Only LSTM can provide some fluctuation in the prediction, but the variation is within a very small range like 1 in this example. It makes sense that all the predictions are within a small range considering the overall range of the true views is between 0 and 5.

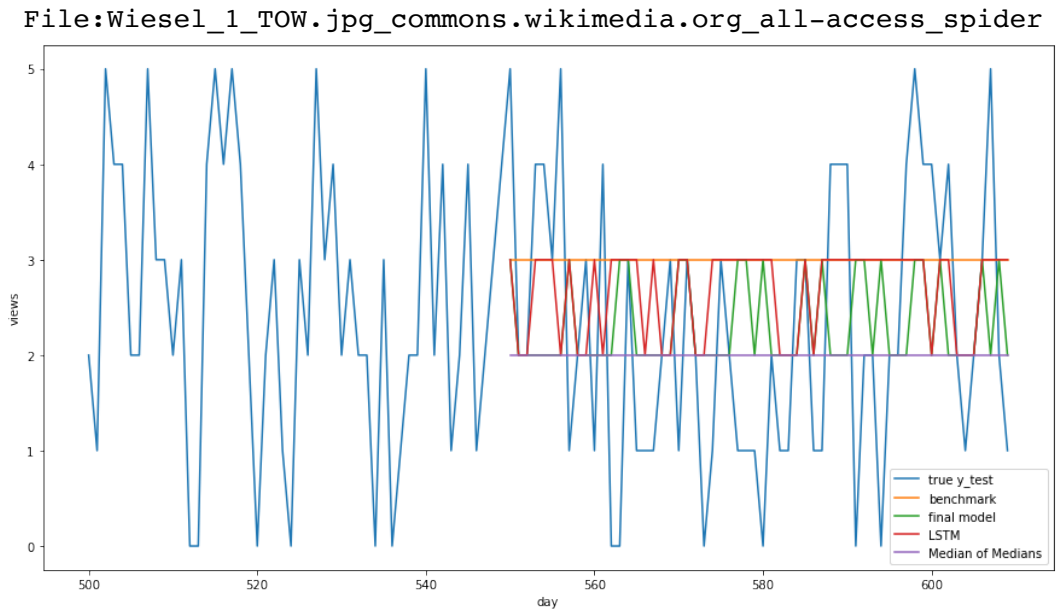

**Figure 9**. True values and predictions from different methods

## Conclusion

**Free-Form Visualization**

The discussion with plots in last section illustrates that our model can capture the seasonality and some monotone in the time series. Let's plot the results for the 4 pages shown in Figure 1, and further discuss the performance of our final model in general.

For the $1^{st}$ page (left-upper) and $4^{th}$ page (right-bottom), most of the views are small except some high spikes. The predictions for this type of series has little variation, so the benchmark and our model are very close while the prediction from final model is slight better in variation. For the $2^{nd}$ page (right-upper), the daily views have some seasonality with monotonic trend. The predictions from our Model can capture the fluctuations. Similarly, the $3^{rd}$ page (left-bottom) has seasonality with monotone, but in addition, there is an extremely high spike, see Figure 1. The predictions from final model is much better than that from benchmark.

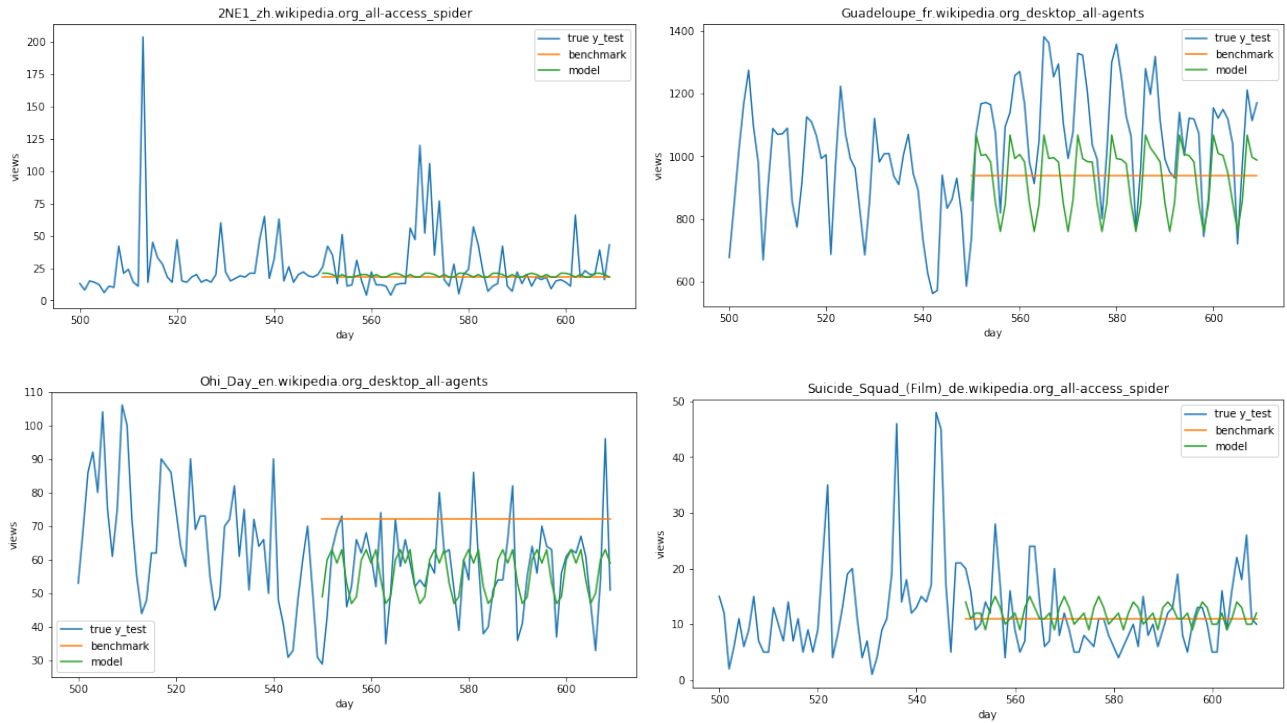

**Figure 10**. Comparison of Final Model and Benchmark Model

**Reflection**

Many time series just follow random walks and are not stationary, so it is difficult to find patterns within it and the reason causing fluctuations behind it. In this specific problem, we need to predict a future sequence which consists of not just one step but 60 steps. The error will cumulate if we forecast the future step one by one. What makes it more complicated as well as interesting is to predict more than 145K time series at once. Therefore, I would like to regard this problem as a multiple multistep time series forecasting.

The variation of different time series and the huge volume of it make traditional ARIMA model or other linear models fail to return a good prediction quickly and accurately. Deep neural network plays well in this field with its capability in capturing nonlinear patterns, and the huge training set becomes a help to generate a better result. One more advantage of deep learning is that it saves us time and pain in doing feature engineering.

The final model with simply implemented LSTM architecture returns a reasonable and good prediction in general. With further parameter tuning, the prediction can be better. However, there are still a lot of space to improve.

**Improvement**

As the problem is to predict sequence from a historical sequence, a better way is to use two long-short term memory networks where one is encoder to take historical sequence and the other is decoder to translate it into a future sequence. This kind of model is called Seq2Seq model (see [7] for one example).

In addition to enhance the algorithms, we can also improve the prediction by preprocessing the initial time series or engineering more features so that to describe the trends of the data. So far many of the time series are not stationary. We can differentiate the sequences and make prediction on the difference. By introducing more features, we can define medians of different lags and build neural networks on them.

**Acknowledgment**

Many thanks to Kaggle for holding this challenging and interesting competition. Many thanks to Udacity Support Team for the capstone project template, reviews and suggestion.